\documentclass{article} 
\usepackage{iclr2026_conference,times}

\usepackage{amsmath,amsfonts,bm}

\def\eqref#1{equation~\ref{#1}}

\def\1{\bm{1}}

\DeclareMathAlphabet{\mathsfit}{\encodingdefault}{\sfdefault}{m}{sl}
\SetMathAlphabet{\mathsfit}{bold}{\encodingdefault}{\sfdefault}{bx}{n}

\usepackage{hyperref}
\usepackage{url}
\usepackage{graphicx}
\usepackage{listings}
\usepackage{subfig}
\usepackage{float}
\usepackage[margin=1in]{geometry}
\usepackage[utf8]{inputenc}
\usepackage{multirow}
\usepackage{multicol}
\usepackage[most]{tcolorbox}
\usepackage{fontawesome5} 
\usepackage[table]{xcolor}
\usepackage{colortbl}

\usepackage{algorithm}
\usepackage{algpseudocode}
\usepackage{amsmath}
\usepackage{amssymb}
\usepackage{mathtools}
\usepackage{float}
\usepackage{wrapfig}
\usepackage{booktabs}
\definecolor{codebg}{rgb}{0.97,0.97,0.97}
\definecolor{keyword}{RGB}{0,0,180}
\definecolor{string}{RGB}{170,0,0}
\definecolor{comment}{RGB}{0,128,0}

\definecolor{myorange}{HTML}{CC6600}
\definecolor{mydarkgreen}{HTML}{008000} 
\definecolor{myred}{HTML}{0000E0}

\definecolor{codeblue}{RGB}{0, 102, 204}
\definecolor{codegreen}{RGB}{0, 128, 0}
\definecolor{codegray}{RGB}{128, 128, 128}
\definecolor{backcolor}{RGB}{248, 249, 250}
\definecolor{accentcolor}{RGB}{59, 130, 246}

\tcbset{
  highlightstyle/.style={
    enhanced,
    sharp corners,
    colback=gray!5,
    colframe=black!60,
    boxrule=0.4pt,
    arc=2mm,
    left=4mm,
    right=4mm,
    top=2mm,
    bottom=2mm,
    fonttitle=\bfseries,
    coltitle=black,
    title style={left color=gray!10!white,right color=gray!5!white},
  }
}
\newtcolorbox{prompttemplate}{
    colback=backcolor,
    colframe=accentcolor,
    boxrule=1pt,
    arc=5pt,
    left=10pt,
    right=10pt,
    top=10pt,
    bottom=10pt,
    fonttitle=\bfseries\large,
    title=Claim generation prompt template
}

\newtcolorbox{verificationtemplate}{
    colback=backcolor,
    colframe=accentcolor,
    boxrule=1pt,
    arc=5pt,
    left=10pt,
    right=10pt,
    top=10pt,
    bottom=10pt,
    fonttitle=\bfseries\large,
    title=Claim verification prompt template
}

\lstdefinestyle{pythonic}{
    language=Python,
    backgroundcolor=\color{codebg},
    basicstyle=\ttfamily\footnotesize,
    keywordstyle=\color{keyword}\bfseries,
    stringstyle=\color{string},
    commentstyle=\color{comment}\itshape,
    showstringspaces=false,
    frame=lines,
    breaklines=true,
    numbers=left,
    numberstyle=\tiny\color{gray},
}

\title{Probing the effectiveness of World Models for Spatial Reasoning through Test-time Scaling}

\author{
\centerline{Saurav Jha\textsuperscript{1,2}\thanks{Correspondence to: saurav.jha@mila.quebec}
 \quad
M.~Jehanzeb Mirza\textsuperscript{3} \quad
Wei Lin\textsuperscript{4} \quad
Shiqi Yang\textsuperscript{5} \quad
Sarath Chandar\textsuperscript{1,2,6}}\\
\centerline{\textsuperscript{1}MILA -- Quebec AI Institute} \\
\centerline{\textsuperscript{2} Polytechnique Montréal} \\
\centerline{\textsuperscript{3}MIT CSAIL} \\
\centerline{\textsuperscript{4}Institute for Machine Learning, Johannes Kepler University Linz} \\
\centerline{\textsuperscript{5}Nankai University} \\
\centerline{\textsuperscript{6}Canada CIFAR AI Chair}
}
\iclrfinalcopy 
\begin{document}

\maketitle

\begin{abstract}
Vision–Language Models (VLMs) remain limited in spatial reasoning tasks that require multi-view understanding and embodied perspective shifts. Recent approaches such as MindJourney \citep{Yang2025MindJourneyTS} attempt to mitigate this gap through test-time scaling where a world model imagines action-conditioned trajectories and a heuristic verifier selects helpful views from such trajectories.
In this work, we systematically examine how such test-time verifiers behave across benchmarks, uncovering both their promise and their pitfalls.
Our uncertainty-based analyses show that MindJourney’s verifier provides little meaningful calibration, and random scoring often reduces answer entropy equally well, exposing systematic action biases and unreliable reward signals.
To mitigate these, we introduce a Verification through Spatial Assertions (ViSA) framework that grounds the test-time reward in verifiable, frame-anchored micro-claims. This principled verifier consistently improves spatial reasoning on the SAT-Real benchmark and corrects trajectory-selection biases through more balanced exploratory behavior.
However, on the challenging MMSI-Bench, none of the verifiers, including ours, achieve consistent scaling, suggesting that current world models form an information bottleneck where imagined views fail to enrich fine-grained reasoning.
Together, these findings chart the bad, good, and ugly aspects of test-time verification for world-model-based reasoning. Our code is available at \url{https://github.com/chandar-lab/visa-for-mindjourney}.
\end{abstract}

\section{Introduction}

Spatial reasoning—the ability to infer 3D structure, object relations, and transformations across viewpoints—remains a persistent gap in Vision–Language Models (VLMs) \citep{yang2025thinking, wang2024is, chen2025why}.  MindJourney (MJ) \citep{Yang2025MindJourneyTS} attempts to close this gap through test-time scaling with world models, where imagined trajectories over actions are generated and scored by heuristic “helpfulness” verification. Yet, the faithfulness of such verifier remains unclear. We begin by probing how much does MJ’s verifier effect the answer selection confidence of a VLM to examine whether it genuinely helps with spatial reasoning. Entropy-based ablations reveal that its scoring mechanism barely reduces uncertainty compared to random selection, often reinforcing systematic biases in action selection. These findings expose a lack of calibration in heuristic verifiers that naively exploit the design of existing VLMs in considering only the global image context for  reasoning \citep{cheng2024spatialrgpt}.

\begin{figure}[!t]
    \centering
    \includegraphics[width=.8\textwidth, trim = 0cm 13.5cm 0cm 0cm, clip]{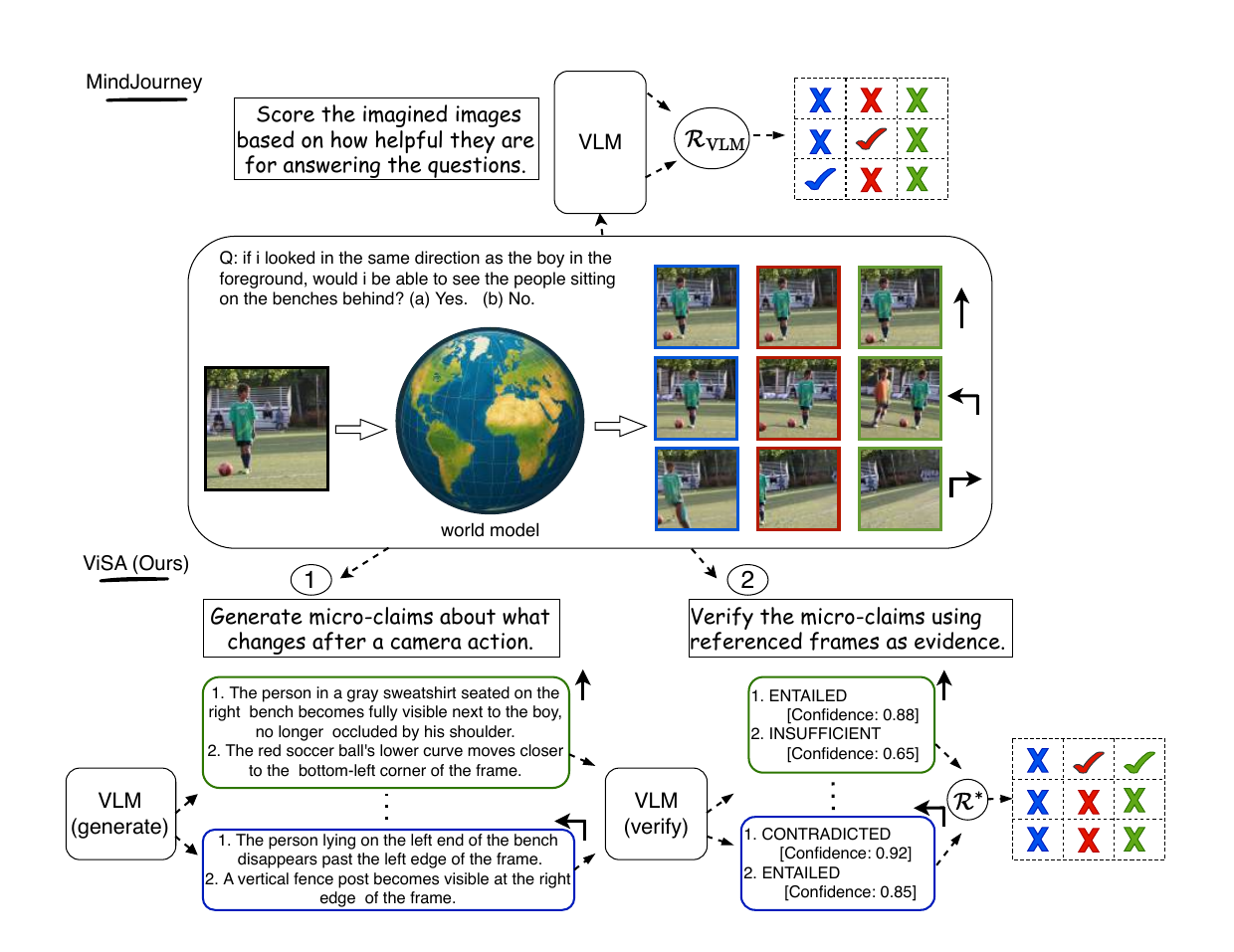}
        \vspace{-0.2in}
    \caption{Illustration of the pipelines for accumulating the evidence buffers (marked by 3x3 boxes) in MindJourney (MJ) vs. our ViSA. While MJ directly prompts a VLM to score all world model views jointly, ViSA does so at a more granular frame-level where a claim generator VLM is first asked to generate frame-anchored micro-claims about the observable changes due to an action. This is followed by a claim verifier VLM being asked to evaluate these claims using the same anchored frames as evidence. Based on the evaluation results, a test-time reward ($\mathcal{R}^*$) then scores the individual frames. \textbf{Arrows} denote egocentric actions including moving forward ($\uparrow$), turning left ($\longleftarrow$), and turning right ($\longrightarrow$). \textbf{Colormap:} \textcolor{blue}{blue}, \textcolor{red}{red} and \textcolor{green!45!black}{green} denote increasing order of magnitudes for each action.}
    \label{fig:whole_pipeline}
\end{figure}

To address these shortcomings, we look at test-time verification through the lens of \textit{proposer-solver} paradigm, which has recently been traction due to its ability to help VLMs  generate their own diverse and challenging reward signals while reducing their reliance on costly verification \citep{thawakar2025evolmm, he2025visplay}. Namely, we introduce Verification through Spatial Assertion (ViSA), a claim-based assertion framework that grounds trajectory evaluation in explicit, frame-level regional information for reasoning (see fig. \ref{fig:whole_pipeline}). Instead of relying on black-box scalar rewards, ViSA prompts the model to propose micro-claims describing spatial relations observed in imagined frames, and evaluates them for consistency and informativeness. Based on the evaluations, each imagined view from the world model is assigned an evidence quality (EQ) score as a principled test-time reward that reflects the frame's relevance to the question alongside the verifier VLM's overall confidence in its evaluations. In contrast to prior proposer–solver methods, which are used almost exclusively for training-time self-improvement \citep{zhao2025absolute}, ViSA is the first such technique to apply a proposer–solver interaction at test time for compute scaling in spatial reasoning tasks for VLMs.

Our experiments show that ViSA  achieves a significant performance gain on the SAT-Real benchmark \citep{ray2025sat} while yielding more balanced exploration behavior over MJ's heuristic approach.  However, when extended to MMSI-Bench \citep{yang2025mmsi}, which emphasizes fine-grained relational and attribute reasoning, all verifiers, including ours, plateau and fail to scale with additional imagined views. Ablating the perceptual quality scores of the generated views point to an underlying information bottleneck in using current world models for test-time scaling of VLMs: when simulated trajectories fail to introduce new or reliable spatial cues, verification alone cannot help. Together, these findings offer a nuanced view of test-time verification—clarifying when it works, why it fails, and what future world model-based approaches must overcome for robust spatial reasoning.

\section{Methods}

\subsection{Notation and Problem Formulation}

Let $\mathbf{x}_0$ denote the initial input image depicting a 3D scene, and $q$ represent a spatial reasoning question with answer choices $\mathcal{A} = \{\alpha_1, \ldots, \alpha_n\}$. Our goal is to predict the correct answer $\alpha^* \in \mathcal{A}$. Traditional VLMs model this as $P(\alpha^* | \mathbf{x}_0, q)$, but struggle with complex spatial reasoning requiring \textit{multi-viewpoint} analysis.

\textbf{Test-time verification of world models' outputs:} At inference-time, a pre-trained video-diffusion world model $\mathcal{W}$ acts as a markov decision process  \citep{cong2025can} when prompted with the reference image $\mathbf{x}_0$, a prompt $c$, and a trajectory $\tau = (f_1, \ldots, f_{t-1})$ comprising a sequence of frames $f_i$ to autoregressively generate an imagined video:
\begin{equation}
   \mathbf{V}_t = \mathcal{W}(\mathbf{V}_t|\mathbf{x}_0, c, \tau_{1:t-1}),  
    \label{eq:world_model_gen}
\end{equation}
    where $\mathbf{V}_t$ is composed of $m$ frames $(\mathbf{x}_1, \ldots, \mathbf{x}_m)$ that can be readily used as helpful signals for a downstream  task. However, not all these imagined frames in $\mathbf{V}_t$ are useful for the task's performance. Hence, a test-time reward function $\mathcal{R}$ helps quantify the relevance of the generated frames for a given task  \citep{cong2025can}.

\textbf{MindJourney's heuristic test-time verification:} MindJourney (MJ) \citep{Yang2025MindJourneyTS} employs the imagined frames of the world model as \textit{egocentric rollouts} to address spatial reasoning through beam search, \textit{i.e.,} the output $\mathbf{V}_t$ simulates how an embodied agent would navigate the static 3D scene referenced by the input image $\mathbf{x}_0$. For this, an action space of a small set of primitive actions $a_i$ is used, namely $a_i \in $ \{move-forward $d$, turn-left $\theta_l$, turn-right $\theta_r$\}, where $d$ and $(\theta_l, \theta_r)$ are the actions' magnitudes in meters and degrees, respectively. The frames $f_i$ comprising the  trajectory input $\tau$ to the world model (\eqref{eq:world_model_gen}) are effective mappings of the action $a_i$ by a camera-pose transformation: $\psi(a_i) = f_i \in \text{SE}(3)$. As its test-time reward function, MJ uses a \textit{helpfulness} beam search, where each candidate frame $\mathbf{x}_i \in \mathbf{V}_t$  is assigned a scalar reward score $\mathcal{R}_{\text{VLM}}(\mathbf{V}_t, q)$ by a VLM. The top-$k$ ranked frames in $\mathbf{V}_t$ are stored in an evidence buffer $\varepsilon$ comprising the global observation set of helpful frames seen until a beam node of given depth $\gamma$. Spatial reasoning is thus posed using the evidence buffer as $P(\alpha^* | \mathbf{x}_0, q, \epsilon)$.

\begin{wrapfigure}{r}{0.35\textwidth}
        \centering
                \vspace{-0.1in} 
        \includegraphics[width=\linewidth]{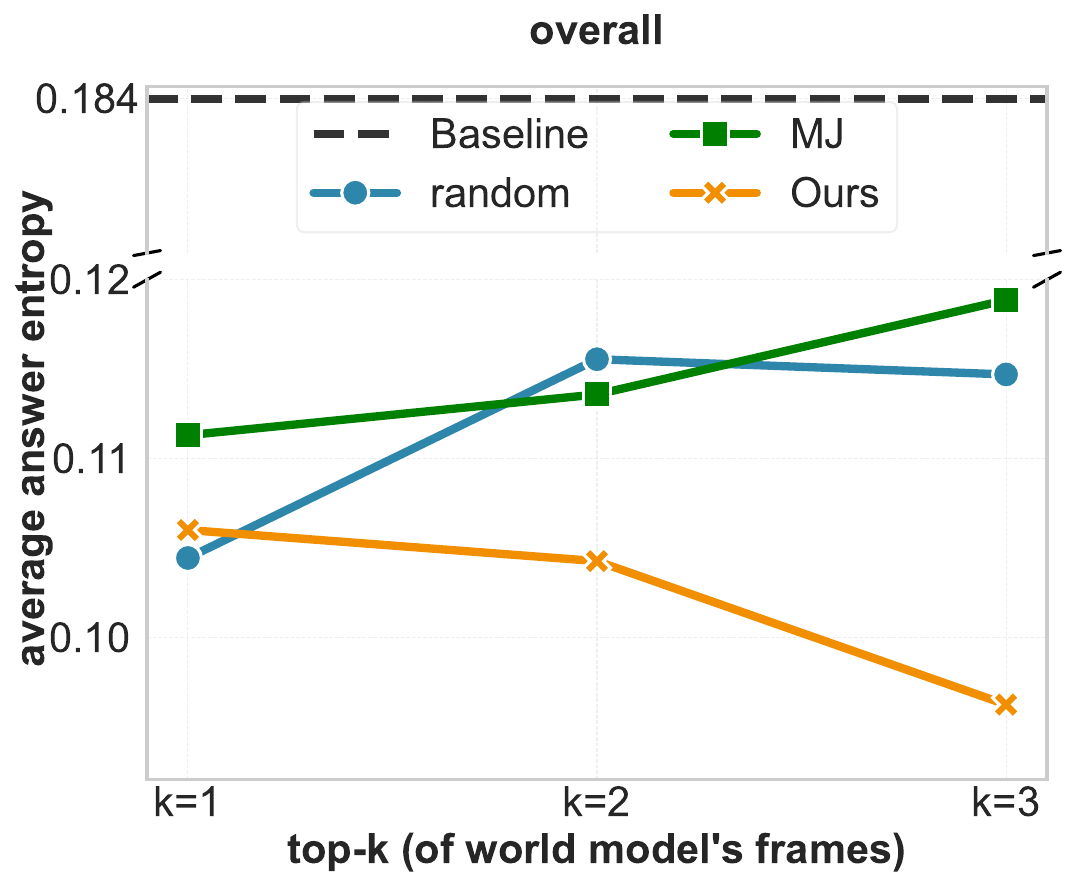}
        \caption{Average entropy of InternVL3-14B over the answers of 50 randomly sampled questions from SAT-Real.}
        \label{fig:avg_entropy}
        \vspace{-0.1in}
\end{wrapfigure}
\textbf{How reliable is MindJourney's verifier?} We hypothesize that for MJ's test-time scaling to be effective, its \textit{helpfulness} score $\mathcal{R}_{\text{VLM}}$ should, in general,  reflect improved reasoning confidence as it grows more permissive, \textit{i.e.}, $\mathcal{R}_{\text{VLM}}$ with a larger top-$k$ should help boost the VLM's confidence in question answering with more qualitative evidence. Subsequently, we study how MJ’s verifier influences the VLM’s entropy over answer choices by comparing it against a \textit{random} counterpart (see Sec. \ref{ref:uncertainty_algo}). 
Fig. \ref{fig:avg_entropy} shows the average entropy scores of the models in choosing the answer choices. A baseline (without test-time scaling) clearly exhibits higher entropy for answer selection. While for all top-$k$, both the test-time verifiers reduce the uncertainty over the baseline, random selection surprisingly produces lower entropy, and hence better calibration than MJ's helpfulness scoring. We further analyze correct versus incorrect predictions where for correct answers, a well-calibrated verifier should help provide more informative evidence with larger top-$k$ while for wrong answers, more evidence should reveal inconsistencies that make the model less confident. Fig. \ref{fig:overall_entropy_trend}  shows that for correct answers, random selection again yields lower entropy than MJ for a more permissive top-$k$.  For wrong answers, MJ's confidence shows a mixed trend while random scoring leads to a consistent increase in the VLM's uncertainty with larger top-$k$. These results suggest that MJ’s helpfulness reward $\mathcal{R}_{\text{VLM}}(\mathbf{V}t, q)$, though capable of pruning implausible trajectories, is fundamentally misaligned with reasoning quality. This could be attributed to the verifier VLM jointly ranking multiple imagined views which makes its scores  highly sensitive to the number and similarity of candidate frames. Visually similar inputs thus tend to receive near-identical values. Subsequently, with a lack of explicit incentive for fine-grained evaluations, MJ's verifier VLM might tend to reward visually salient or novel frames rather than those truly informative for reasoning \citep{dahou2025salbench, vo2025vision, campbell2024understanding}. This calls for a more principled verifier $\mathcal{R}^*(\mathbf{V}_t, q)$ grounded in measurable informativeness.

\subsection{Verification through Spatial Assertions (ViSA)}
To address the flaw in MindJourney’s helpfulness reward $\mathcal{R}_{\text{VLM}}(\mathbf{V}_t, q)$, we propose VerIfication through Spatial Assertions (ViSA) -- a two-stage verification framework (see fig. \ref{fig:verifier}) that grounds test-time reasoning in fine-grained, verifiable spatial assertions. ViSA decomposes the evaluation of imagined frames into two interpretable steps: \textit{claim generation} and \textit{claim verification}. In the first stage, for each candidate frame $\mathbf{x}_i \in \mathbf{V}_t$, the VLM generates a set of question-conditioned micro-claims $\mathcal{C} = \{c_1, \ldots, c_k\}$ describing spatial relations, object attributes, or dynamic changes with respect to the input image $\mathbf{x}_0$. Each claim explicitly captures a localized visual hypothesis (\textit{e.g.}, “the red cube moved behind the sphere”) rather than a global saliency cue. In the second stage, every claim is verified against its corresponding frame by a dedicated verifier, which produces a verdict $v_j \in \{\text{ENTAILED}, \text{CONTRADICTED}, \text{INSUFFICIENT}\}$ and a confidence score $conf_j \in [0,1]$. This explicit decomposition encourages the VLM to attend to distinct spatial regions and separate question-relevant evidence from irrelevant context, while also producing interpretable reasoning traces that expose which claims supported the final decision.\footnote{This decomposition  separates \textit{reasoning} from \textit{validation}: a single VLM performing both can conflate description with judgment, thus reinforcing its biases. ViSA ensures that claims about a scene are produced independently and then objectively tested.}
To translate the verified micro-claims into a frame-level score, we define an evidence quality (EQ) score that aggregates the verifier’s outputs across all claims for a frame $\mathbf{x}_i$:
\begin{equation}
\text{EQ}(\mathbf{x}_i, q)
= 
\underbrace{
\left(
\frac{1}{|\mathcal{C}|} \sum_{j=1}^{|\mathcal{C}|} \mathbb{1}[v_j = \text{ENTAILED}]
\right)
}_{\text{Proportion of entailed claims}}
\times
\underbrace{
\left(
\frac{1}{|\mathcal{C}|} \sum_{j} conf_j
\right).
}_{\text{Average confidence over all claims}}
\end{equation}
By balancing coverage (how many claims are entailed) with certainty (how confidently they are verified), $\text{EQ}$  rewards frames that not only yield a high fraction of verifiable micro-claims but also do so with strong verifier confidence. 
$\text{EQ}$ can thus be seen as a \textit{soft ensemble} over multiple localized evidential checks within each frame, aggregating per-claim confidences to smooth out noisy judgments. Since these confidence scores are subjective to the VLM’s internal calibration, averaging across multiple claims helps us stabilize the overall reliability of frame-level informativeness.

\section{Results}

\textbf{Dataset and evaluation framework.} 
We evaluate our proposed verifier on two benchmarks: (a) \textbf{SAT} \citep{ray2025sat}, containing 5 spatial reasoning categories, and (b) a subset of \textbf{MMSI-Bench} \citep{yang2025mmsi}, which includes 11 fine-grained spatial reasoning categories. 
We use InternVL3-14B as the backbone VLM, and Stable Virtual Camera~\citep{zhou2025stable} as our world model. All experiments are conducted on 2$\times$A100~80GB GPUs. For beam nodes of depth $\gamma = 2$, we ensure fair comparisons by retaining the \textit{exploration} beam search of MJ \citep{Yang2025MindJourneyTS}.

\subsection{Results on SAT-Real}
SAT-Real split covers 150 real-image questions that test spatial reasoning skills on indoor and outdoor scenes. Table~\ref{tab:satreal} summarizes accuracy for different sizes of evidence buffer (top-$k$) on beam nodes of various depths $(\gamma \in \{1,2\})$. Both random and MJ-based selection improve significantly over the baseline InternVL3-14B model, confirming that world-model rollouts offer valuable auxiliary evidence for spatial reasoning. However, our proposed ViSA achieves the best overall accuracy across all top-$k$ settings (65.3–72.7\%), consistently outperforming random and MJ. Notably, for beam depth of 1 $(\gamma = 1)$, the random verifier remains surprisingly competitive, suggesting that even unstructured exploration contributes useful diversity. However, our principled verification approach further amplifies this benefit through question-specific claim consistency checking, and achieves best results across all but the Perspective questions.
\begin{table}[t]
\scriptsize
\setlength{\tabcolsep}{2.5pt}
\renewcommand{\arraystretch}{1.2}
\centering
\caption{
\textbf{SAT-Real results.} Accuracy of different test-time verifiers for top-$k \in \{1,2,3,4\}$ selected frames, separated by beam node depth $\gamma \in \{1,2\}$. 
\textbf{Abbreviations:} EgoM=Ego Motion, ObjM=Object Motion, EgoAct=Ego Action, GoalAim=Goal Aiming, Pers=Perspective. Best results across columns are in \textbf{bold}.
}
\label{tab:satreal}
\begin{tabular}{@{}l|c|ccccc|c|ccccc@{}}
\toprule 
\rowcolor{blue!8}
& \multicolumn{6}{c|}{\textbf{Beam depth $\gamma=1$}} & \multicolumn{6}{c}{\textbf{Beam depth $\gamma=2$}} \\
\cmidrule(lr){2-7} \cmidrule(lr){8-13}
\textbf{Model} & \textbf{Avg.} & \textbf{EgoM} & \textbf{ObjM} & \textbf{EgoAct} & \textbf{GoalAim} & \textbf{Pers} & \textbf{Avg.} & \textbf{EgoM} & \textbf{ObjM} & \textbf{EgoAct} & \textbf{GoalAim} & \textbf{Pers} \\ 
\midrule
\text{InternVL3-14B (baseline)} & 41.33 & 52.17 & 69.57 & 32.43 & 32.35 & 33.33 & 41.33 & 52.17 & 69.57 & 32.43 & 32.35 & 33.33 \\
\midrule
Random ($k=1$) & 63.33 & 65.22 & 56.52 & 78.38 & 70.59 & 42.42 & 61.33 & 52.17 & 60.87 & 70.27 & 79.41 & 39.40 \\
Random ($k=2$) & 66.00 & 73.91 & 56.52 & 70.27 & 76.47 & 51.52 & 58.67 & 52.17 & 52.17 & 59.46 & 76.47 & 48.48 \\ 
Random ($k=3$) & 64.00 & 82.61 & 56.52 & 62.16 & 76.47 & 45.45 & 59.33 & 65.22 & 52.17 & 67.57 & 70.59 & 39.40 \\ 
Random ($k=4$) & 63.33 & 69.57 & 60.87 & 54.05 & 76.47 & \textbf{57.58} & 67.33 & 73.91 & 56.52 & 78.38 & 73.53 & 51.52 \\ 
\midrule
MJ ($k=1$) & 63.33 & 65.22 & 52.17 & 72.97 & 76.47 & 45.45 & 66.67 & 69.57 & 69.57 & 70.27 & 79.41 & 45.45 \\ 
MJ ($k=2$) & 67.33 & 60.87 & 65.22 & 75.68 & 79.41 & 51.52 & 70.67 & 73.91 & \textbf{73.91} & 75.68 & 76.47 & 54.55 \\ 
MJ ($k=3$) & 64.67 & 73.91 & 65.22 & 64.86 & 73.53 & 48.48 & 63.33 & 52.17 & 65.21 & 64.86 & 76.47 & 54.55 \\
MJ ($k=4$) & 64.67 & 73.91 & 65.22 & 64.87 & 76.47 & 45.45 & 66.67 & 69.57 & 65.21 & 72.97 & 79.41 & 45.45 \\
\midrule
Ours ($k=1$) & 65.33 & 65.22 & 65.22 & 72.97 & 79.41 & 42.42 & 69.33 & 73.91 & 65.22 & 81.08 & 70.59 & 54.55 \\ 
Ours ($k=2$) & 68.67 & 69.57 & 69.57 & 75.68 & \textbf{85.29} & 42.42 & 67.33 & 78.26 & 65.21 & 62.16 & 79.41 & 54.55 \\ 
Ours ($k=3$) & 69.33 & 82.61 & 65.21 & 78.38 & 76.47 & 45.45 & 68.67 & 69.57 & \textbf{73.91} & \textbf{81.08} & 70.59 & 48.48 \\ 
Ours ($k=4$) & \textbf{72.67} & \textbf{95.65} & 65.22 & 75.68 & 79.41 & 51.52 & 66.67 & 65.21 & 56.52 & 72.97 & 82.35 & 51.52 \\ 
\bottomrule
\end{tabular}
\vspace{-0.1in}
\end{table}

\begin{figure}[!t]
\centering
    \includegraphics[width=0.9\textwidth]{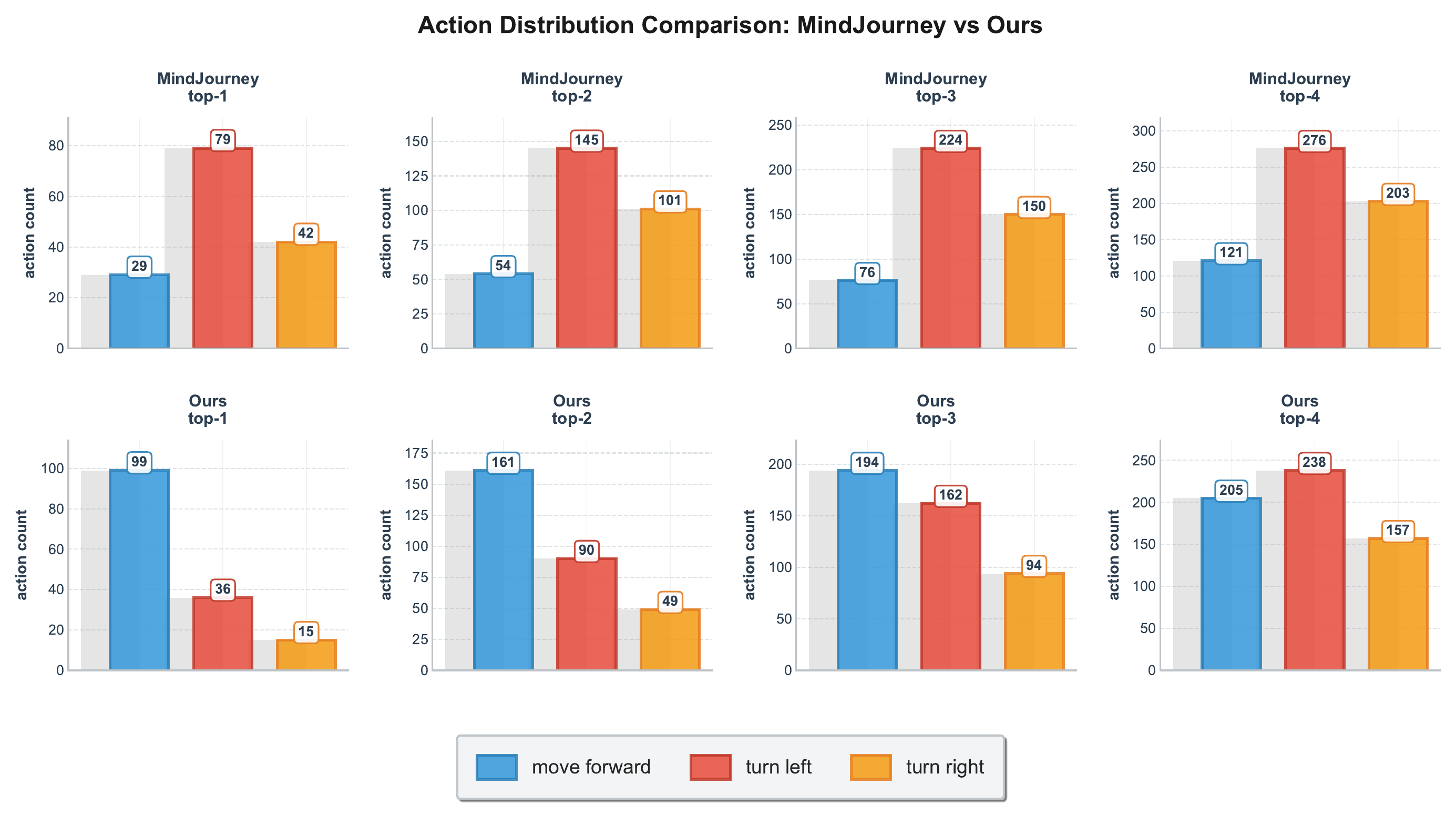}
\caption{Action distribution comparison between MindJourney and ViSA (ours) across different top-$k \in \{1,2,3\}$ values. We consider three action types (move forward, turn left, turn right) for each model's top-$k$ configuration.}
    \label{fig:action_dist_main_paper}
\end{figure}

\textbf{Analyses of selected actions.}
To analyze action selection biases, Figures~\ref{fig:action_dist_main_paper} and~\ref{fig:fine_grained} compare action distributions selected by ViSA (ours) against MindJourney (MJ) on beam node of depth 1. Across all top-$k$ values, MJ exhibits a persistent left-turn bias (46–53\% of all moves) and an overreliance on high-magnitude turns (27° accounting for 62–93 occurrences), indicating suboptimal navigation. In contrast, our verifier yields more balanced and context-aware actions, distributing turns across 9°, 18°, and 27° ranges (22–79, 8–87, and 6–72 occurrences, respectively). These results suggest that our test-time reward formulation not only improves the final accuracy but also regularizes spatial exploration by discouraging redundant or biased trajectories.

\begin{wraptable}{r}{5.5cm}
\vspace{-0.35in}
\centering
\caption{Perceptual quality scores \citep{laion_aesthetics_v1} of world model outputs by dataset and beam depth $\gamma$.}
\label{tab:aesthetic_scores}
\begin{tabular}{lccc}
\toprule
\textbf{Benchmark} & \textbf{$\gamma=1$} & \textbf{$\gamma=2$} & \textbf{Avg.} \\
\midrule
SAT & 5.26 & 4.94 & 5.12 \\
MMSI & 4.61 & 4.42 & 4.53 \\
\bottomrule
\end{tabular}
\vspace{-0.1in}
\end{wraptable}
\subsection{Results on MMSI-Bench}
Table~\ref{tab:mmsi_reduced} reports results on MMSI-Bench, which evaluates a wider set of spatial and relational reasoning types (e.g., Camera–Object, Object–Region) that rely on precise visual cues. While our ViSA shows slight gain on overall accuracy for top-$k=1$, we observe that none of the verifiers show a consistent improvement pattern with increasing top-$k$ or $\gamma$, with performance fluctuating around 27–36\%. This stagnation hints toward an information bottleneck in current world models: imagined views fail to meaningfully alter relational cues, leaving the verifier with little discriminative signal. Comparing the visual fidelity of the generated frames for both benchmarks (using  LAION aesthetic predictor) shows that the generated images on MMSI-Bench exhibit significantly lower perceptual quality across different beam depths (see Table \ref{tab:aesthetic_scores}). Since the test-time verifiers rely on high-level textual (ViSA) or relational (MindJourney) consistency to score imagined frames, these signals are
poorly aligned with the subtle geometric or appearance shifts central to MMSI questions. These findings underscore that while claim-based verification enhances spatial exploration tasks like SAT-Real, its benefits diminish when world-model generations lack the granularity required for relational or attribute-based reasoning. We leave a detailed discussion of these results and implications in App. \ref{app:mmsi-bench-results}.

\section{Conclusion}
This work investigated the effectiveness of test-time verification for spatial reasoning with world models through the lens of the bad, the good, and the ugly. Our analyses revealed that existing heuristic verifiers, such as MindJourney’s helpfulness score, offer limited confidence gains, motivating our claim-based framework ViSA, which grounds reasoning in verifiable spatial assertions and achieves strong improvements on SAT-Real. However, results on MMSI-Bench expose a fundamental information bottleneck, \textit{i.e.}, when imagined evidence lacks physical or relational fidelity, verification offers little benefit. These findings highlight both the promise and the limits of current test-time scaling for VLMs with world models, pointing toward future directions beyond pixel-space planning  \citep{zhang2025can}.

\section*{Acknowledgement}

Saurav Jha is  supported by the IVADO postdoctoral fellowship and the Canada First Research Excellence Fund.
Sarath Chandar is supported by the Canada CIFAR AI Chairs program,
the Canada Research Chair in Lifelong Machine Learning, and the NSERC Discovery Grant. This research was enabled by compute resources provided by Mila.

\bibliography{iclr2026_conference}
\bibliographystyle{iclr2026_conference}
\clearpage
\appendix
\section{Appendix}

\begin{figure}[!h]
    \centering
    \begin{minipage}[t]{0.8\textwidth}
        \centering
                \vspace{0pt} 
        \includegraphics[width=\linewidth]{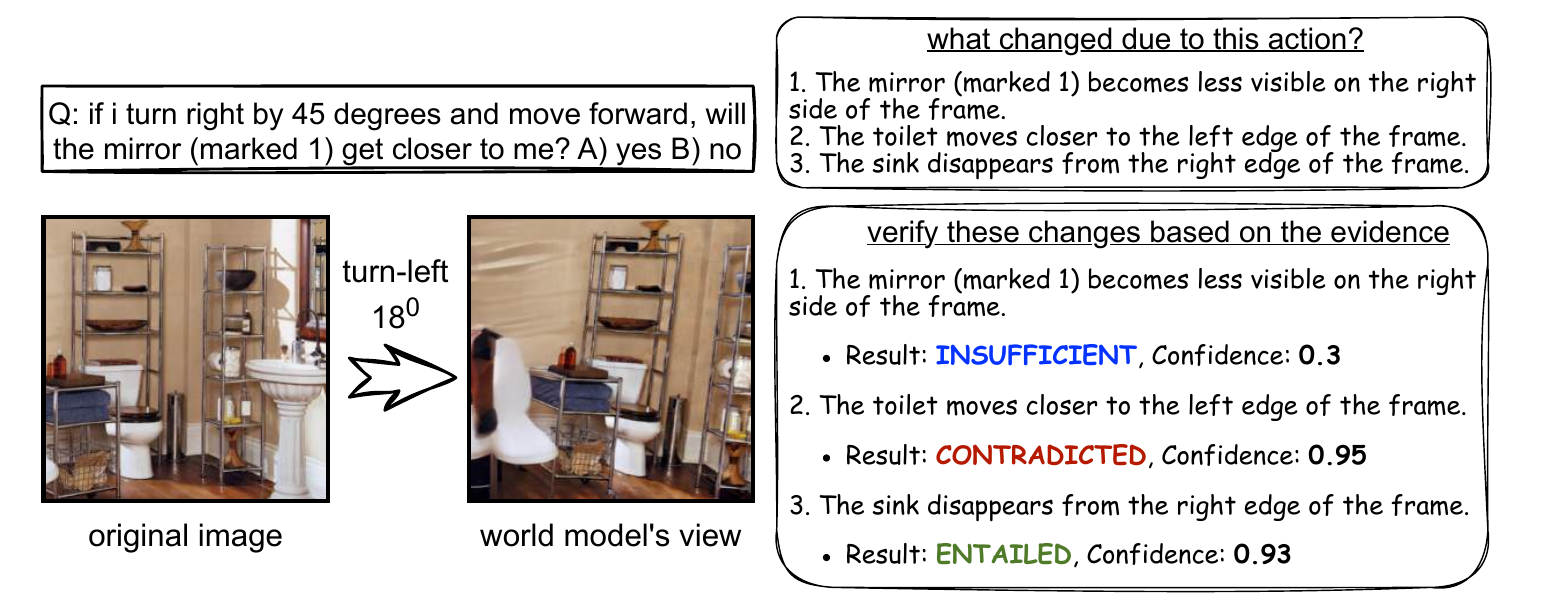}
        \caption{Illustration of claim \textit{generation} and \textit{verification} steps within ViSA.}
        \label{fig:verifier}
        \vspace{-0.2in}
    \end{minipage}
\end{figure}

\subsection{Teacher-forcing for measuring a VLM's uncertainty in Answer Selection}
\label{ref:uncertainty_algo}

To measure a VLM's entropy over answer choices, we aim to reliably record the VLM's log-probabilities over each token of the choices. We thus apply teacher-forcing \citep{agrawal2025uncertainty} across all answer choices, allowing the VLM to output each choice along with its log-probabilities. We then compute the entropy over these probabilities to quantify the answer's uncertainty. Algo. \ref{entropy_algo} depicts this in detail:

\begin{lstlisting}[style=pythonic, caption={Python-styled pseudo-code for computing entropy scores over answer choices using teacher-forcing.}, label={entropy_algo}]
def compute_answer_level_entropy(answer_choices, model, tokenizer, question, images):
    """Compute answer-level entropy from token-level entropies using teacher forcing."""

    answer_log_likelihoods = {}

    for choice in answer_choices:
        full_prompt = question + "\nAnswer:" + choice
        input_ids = tokenizer.encode(full_prompt, return_tensors="pt")

        with torch.no_grad():
            outputs = model.forward(input_ids=input_ids, images=images)
            logits = outputs.logits

        answer_start_idx = find_answer_start_position(input_ids, tokenizer)
        answer_logits = logits[0, answer_start_idx:, :]
        answer_tokens = input_ids[0, answer_start_idx:]

        total_log_likelihood = 0.0
        for i, token_id in enumerate(answer_tokens):
            if i < answer_logits.shape[0]:
                log_prob = log_softmax(answer_logits[i])[token_id]
                total_log_likelihood += log_prob

        answer_log_likelihoods[choice] = total_log_likelihood

    log_likelihoods = list(answer_log_likelihoods.values())
    if not log_likelihoods or any(ll == float('-inf') for ll in log_likelihoods):
        return None

    max_ll = max(log_likelihoods)
    normalized_ll = [ll - max_ll for ll in log_likelihoods]
    probs = softmax(normalized_ll)
    H = -sum(p * log(p) for p in probs if p > 1e-10)
    return H
\end{lstlisting}

\begin{figure}[h!]
    \centering
    \subfloat[All answers]{\includegraphics[width=0.32\textwidth]{figures/entropy_viz/avg_answer_entropy_overall.pdf}\label{fig:sub1}}
    \hfill
    \subfloat[Correct answers]{\includegraphics[width=0.32\textwidth]{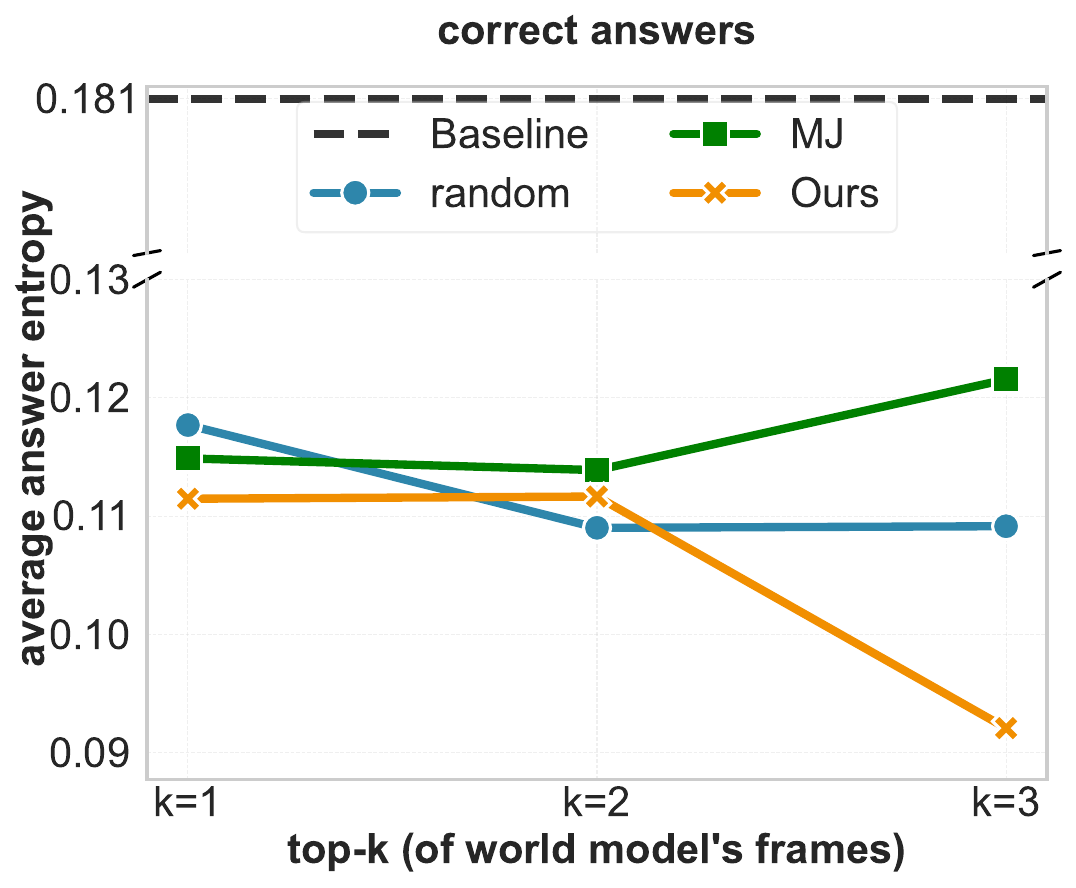}\label{fig:overall_entropy_correct}}
    \hfill
    \subfloat[Wrong answers]{\includegraphics[width=0.32\textwidth]{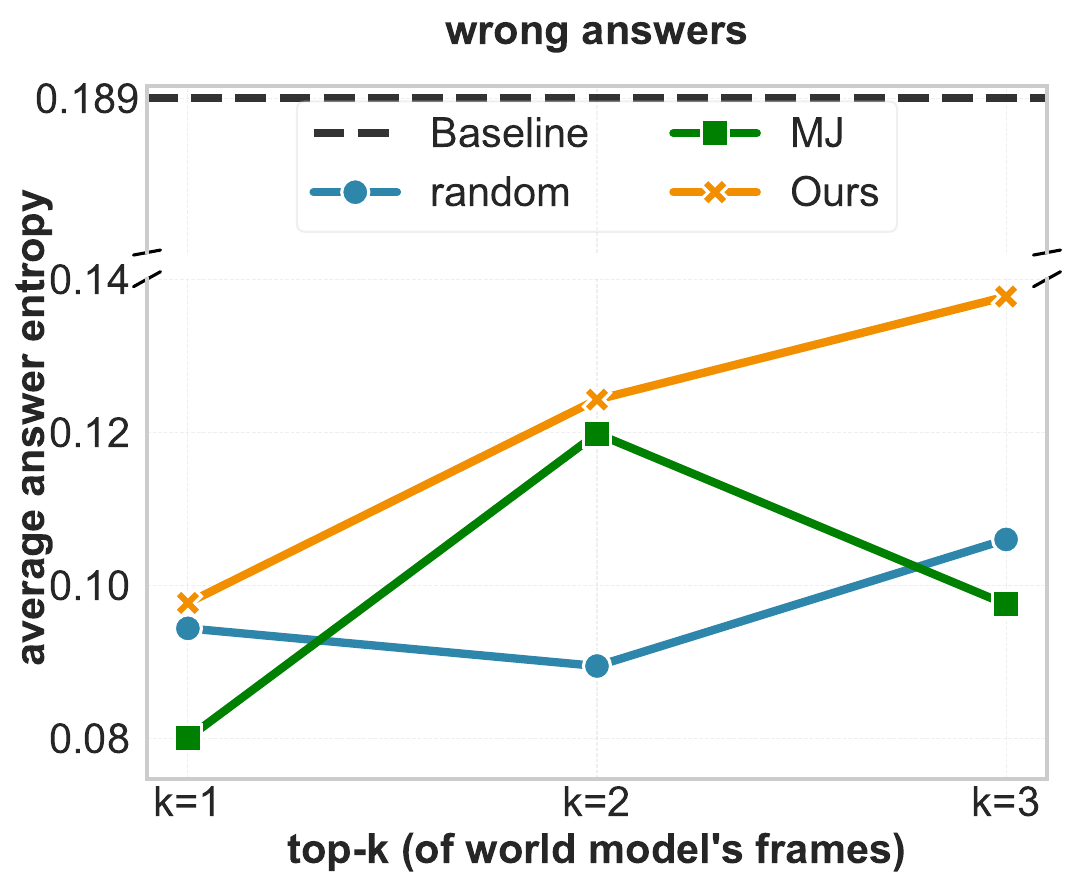}\label{fig:overall_entropy_wrong}}
    \caption{Effect of the verifiers' permissiveness (top-$k$) on answer selection confidence of 50 randomly selected SAT-Real questions grouped by (a) overall, (b) correct answers, and (c) wrong answers. The baseline is InternVL3-14B.}
    \label{fig:overall_entropy_trend}
\end{figure}

\subsection{On datasets and benchmark}

\subsection{SAT-Real results}

\textbf{The SAT-Real test set} represents a comprehensive benchmark for evaluating spatial reasoning capabilities across diverse cognitive tasks, comprising five distinct question categories: EgoM (ego-movement), ObjM (object-movement), EgoAct (ego-action), GoalAim (goal-aiming), and Pers (perspective-taking). This multi-faceted evaluation framework enables systematic assessment of how different action selection strategies impact performance across varying spatial reasoning demands. Table \ref{tab:satreal} compares the results of our verification against MJ and random frame selection for different values of top-$k$. Surprisingly, the test-time scaling approach with a random verifier achieves a substantial improvement ($>20\%$) over the baseline InternVL3-14B model, demonstrating that any additional test-time signal from the world model contains rich information that enables the VLM to significantly enhance downstream spatial reasoning performance. Moreover, the random verifier emerges as a competitive baseline, achieving performance levels (63.33-66.0\%) that closely rival MindJourney's sophisticated action selection (63.33-67.33\%), indicating that systematic action selection may be less critical than previously assumed when world model signals are available. However, our verifier-based approach demonstrates clear advantages over both random and MindJourney baselines, achieving the highest performance across all top-k configurations (65.33-72.67\% avg. accuracy). 

The category-specific analysis reveals ViSA's particular strength in EgoM tasks (95.65\% at top-4 vs. random's 69.57\% and MJ's 73.91\%), which we attribute to the following two reasons: (a) EgoM tasks likely leverage the richest feedback signals from the world model because the model can directly observe the consequences of its navigation decisions, subsequently these may have fewer confounding factors compared to object-movement (where external object dynamics matter) or perspective tasks (where multiple viewpoints must be considered), and (b)  EgoM requires the model to make first-person navigation decisions where systematic biases have the most direct impact - hence our verifier's ability to correct any systematic biases becomes most valuable when the model's own navigation decisions directly determine task success. Lastly, the top-k convergence behavior shows our method's superior scalability - while both random and MindJourney approaches show performance plateaus or degradation with increased candidate diversity, our verifier-based approach demonstrates consistent improvement (65.33\% $\rightarrow$ 72.67\%), indicating robust handling of expanded action spaces and effective mitigation of the systematic biases identified in action distribution analyses.

\begin{figure}[hbp]
\centering

\subfloat[Question type: perspective.]{
    \includegraphics[width=0.7\textwidth]{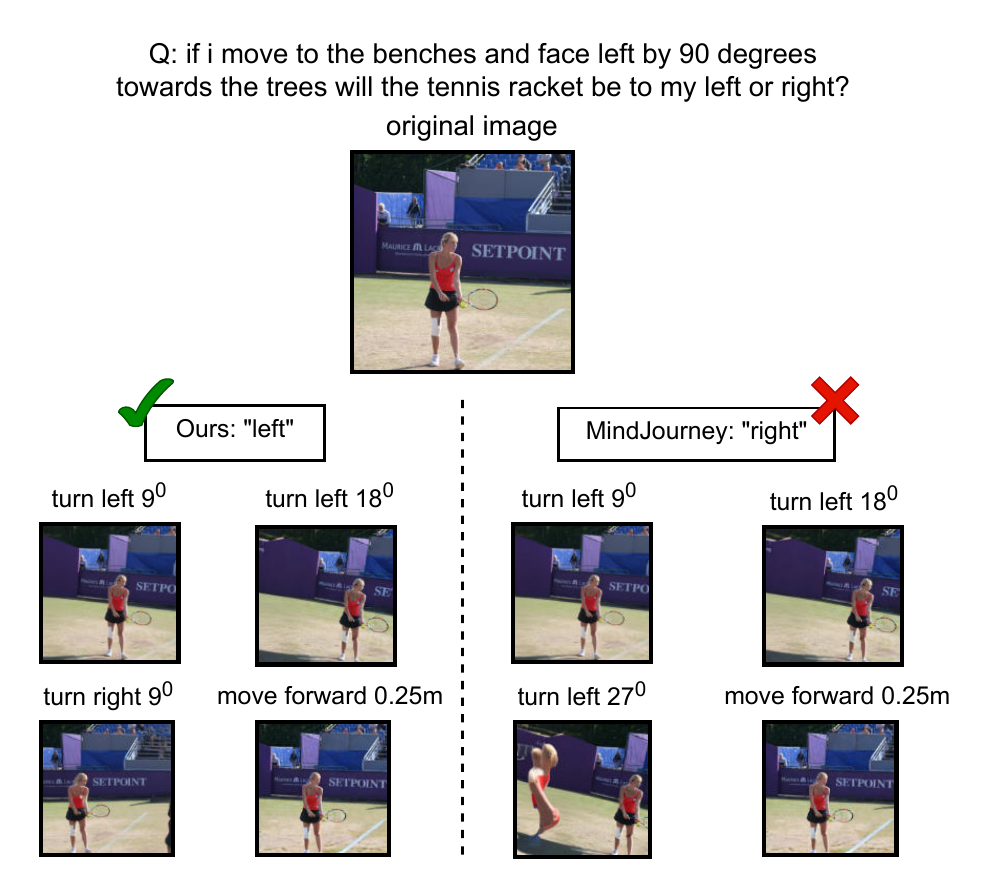}
    \label{fig:example1}
}

\subfloat[Question type: object movement.]{
    \includegraphics[width=0.7\textwidth]{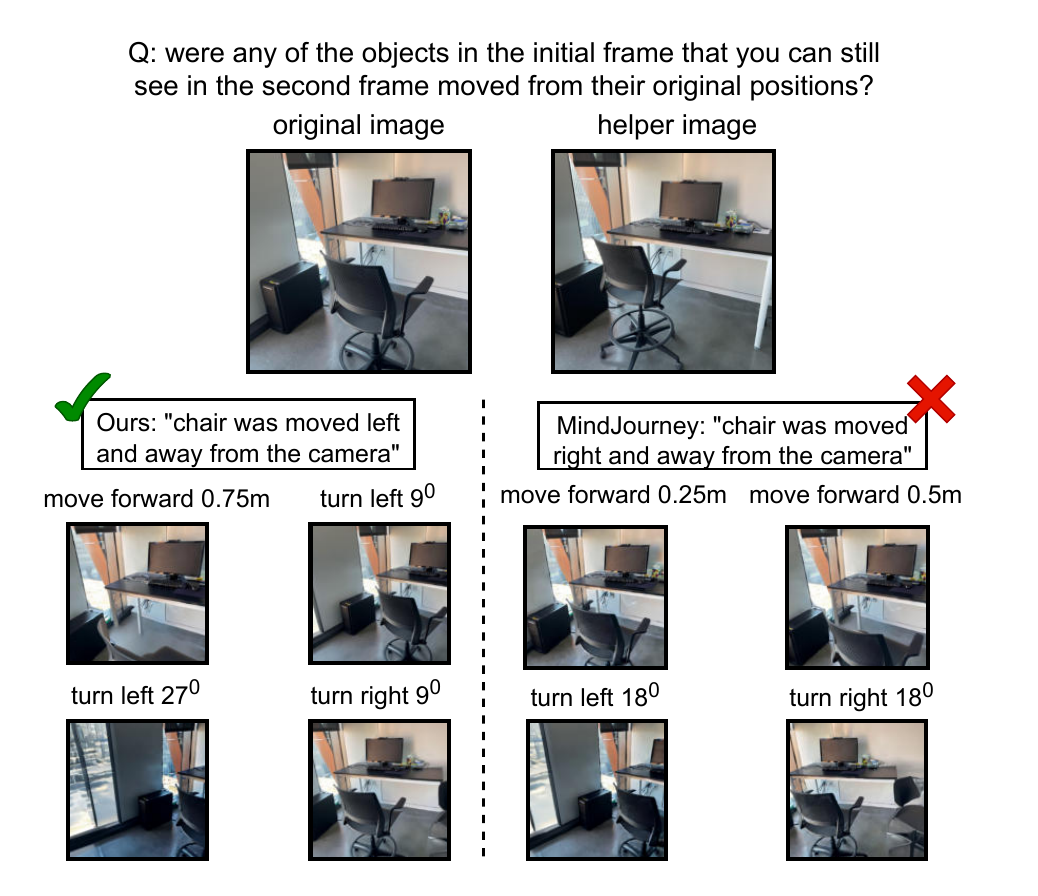}
    \label{fig:example2}
}

\caption{Examples of SAT-Real questions where ViSA (Ours) helps select actions that lead to correct answer while the actions selected by MindJourney's verifier lead to wrong answers. }
\label{fig:examples}

\end{figure}

\subsubsection{Action selection analyses on SAT-Real}
\begin{figure}[!hbp]
\centering

\subfloat[Action distribution comparison between MindJourney and Ours models across different top-k values. The plot shows the frequency of three action types (move forward, turn left, turn right) for each model and top-k configuration.]{
    \includegraphics[width=0.95\textwidth]{figures/sat-real-action-distr/action_distribution_comparison.pdf}
    \label{fig:action_dist}
}

\vspace{0.5cm} 

\subfloat[Fine-grained action distribution analysis showing the magnitude breakdown of each action type. Each subplot displays a 3×3 heatmap where rows represent action types (move forward, turn left, turn right) and columns represent magnitude buckets (0.25m/9°, 0.5m/18°, 0.75m/27°). The color intensity indicates the frequency of each action-magnitude combination.]{
    \includegraphics[width=0.95\textwidth]{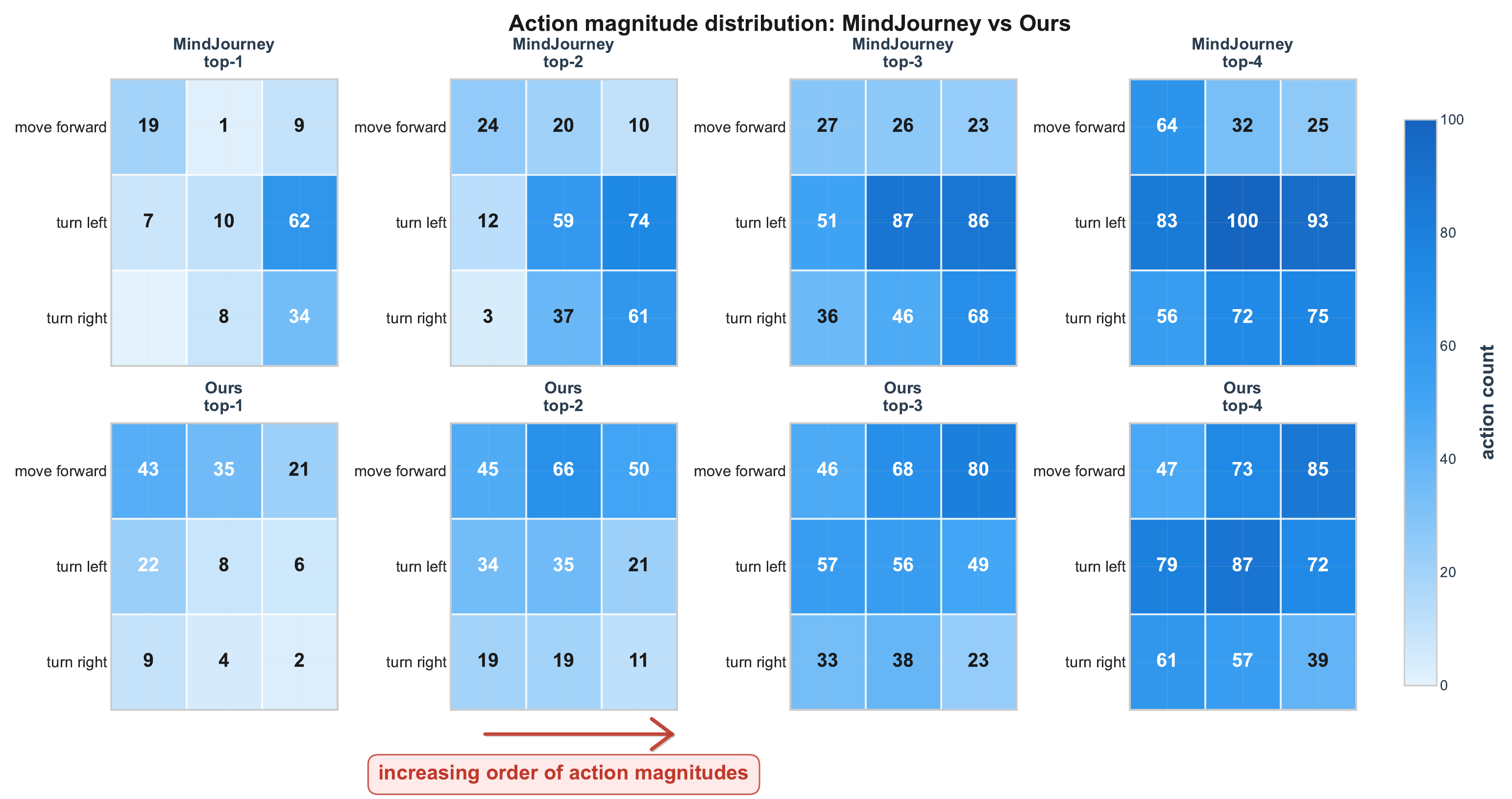}
    \label{fig:fine_grained}
}

\caption{Analyses of SAT-Real action distribution in MindJourney and ViSA (Ours) for beam search depth $\gamma=1$: (a) High-level action type distribution showing the relative frequency of move forward, turn left, and turn right actions across different top-k values. (b) Fine-grained magnitude analysis revealing the specific action magnitudes chosen by each model, providing insights into the precision and consistency of action selection strategies.}
\label{fig:action_analysis}

\end{figure}

Figure \ref{fig:action_analysis} compares the distribution of the actions and sub-actions selected by MindJourney's test-time verifier with ours for different top-$k$ values.
Unlike MindJourney's rigid left-turn preference, our verifier shows adaptive navigation strategies that favor forward movement (34-66\% across top-k values) when appropriate, with more balanced turn distributions (24-40\% left, 10-26\% right). Crucially, our verifier exhibits balanced magnitude selection across all action types - for turns, it distributes choices more evenly across 9° (22-79), 18° (8-87), and 27° (6-72) magnitudes, while for forward movement, it appropriately favors longer 0.75m steps (21-85 occurrences) over shorter 0.25m steps (43-47). This top-k convergence behavior, where the Verifier becomes more exploratory as more candidates are considered (top-1: 66\% forward → top-4: 34\% forward, 40\% left, 26\% right), demonstrates sophisticated decision-making that adapts to the available action space.

\subsection{MMSI-Bench Results}
\label{app:mmsi-bench-results}
\begin{table}[!ht]
\footnotesize
\setlength{\tabcolsep}{3pt}
\renewcommand{\arraystretch}{1.2}
\centering
\caption{
\textbf{Performance on MMSI-Bench.} Accuracy of test-time scaling verifiers on MMSI-Bench across 11 different question categories for top-$k \in \{1,2,3\}$ helpful frames, separated by beam node depth $\gamma \in \{1,2\}$.
\textbf{Abbreviations:} Cam.=Camera, Obj.=Object, Reg.=Region, Meas.=Measurable, Appr.=Appearance, MSR=Multi-Step Reasoning.
}
\label{tab:mmsi_reduced}
\resizebox{\linewidth}{!}{
\begin{tabular}{@{}l|c|cccccc|cc|cc|c@{}}
\toprule
\multirow{2}{*}{\textbf{Model}}
 & \multirow{2}{*}{\textbf{Avg.}}
 & \multicolumn{6}{c|}{\textbf{Positional Relationship}}
 & \multicolumn{2}{c|}{\textbf{Attribute}}
 & \multicolumn{2}{c|}{\textbf{Motion}}
 & \multirow{2}{*}{\textbf{MSR}} \\
\cmidrule(lr){3-8} \cmidrule(lr){9-10} \cmidrule(lr){11-12}
& & \scriptsize Cam–Cam & \scriptsize Obj–Obj & \scriptsize Reg–Reg & \scriptsize Cam–Obj & \scriptsize Obj–Reg & \scriptsize Cam–Reg & \scriptsize Meas. & \scriptsize Appr. & \scriptsize Cam. & \scriptsize Obj. & \\ 
\midrule
InternVL3-14B (baseline) & 27.16 & 13.33 & 20.00 & 33.33 & 33.33 & 33.33 & 20.00 & 33.33 & 13.33 & 26.67 & 53.33 & 16.67 \\ 
\midrule
\rowcolor{blue!8}
\multicolumn{13}{c}{\textbf{Beam depth $\gamma=1$}} \\
\midrule
Random ($k=1$) & 33.33 & 20.00 & 26.67 & 46.67 & 26.67 & 40.00 & 33.33 & 60.00 & 13.33 & 26.67 & 53.33 & 16.67 \\
Random ($k=2$) & 29.63 & 20.00 & 6.67 & 46.67 & 6.67 & 40.00 & 40.00 & 46.67 & 20.00 & 0.00 & \textbf{66.67} & 33.33 \\
Random ($k=3$) & 33.33 & 20.00 & 20.00 & 40.00 & 13.33 & 40.00 & 20.00 & 53.33 & \textbf{26.67} & 20.00 & \textbf{66.67} & \textbf{50.00} \\ \midrule
MJ ($k=1$) & 32.72 & 13.33 & 13.33 & 40.00 & \textbf{46.67} & 46.67 & 33.33 & \textbf{66.67} & 13.33 & 26.67 & 33.33 & 25.00 \\
MJ ($k=2$) & 27.16 & 20.00 & 13.33 & 20.00 & 33.33 & 20.00 & \textbf{46.67} & 40.00 & 13.33 & 13.33 & 53.33 & 25.00 \\
MJ ($k=3$) & 29.63 & 13.33 & 20.00 & 33.33 & 26.67 & \textbf{53.33} & 26.67 & 33.33 & 20.00 & 20.00 & 53.33 & 25.00 \\ \midrule
Ours ($k=1$) & \textbf{35.80} & \textbf{26.67} & 26.67 & 26.67 & 33.33 & 46.67 & 33.33 & \textbf{66.67} & 13.33 & \textbf{33.33} & 53.33 & 33.33 \\
Ours ($k=2$) & 25.93 & 13.33 & 20.00 & 33.33 & 13.33 & 33.33 & 20.00 & 26.67 & 20.00 & 6.67 & \textbf{66.67} & 33.33 \\
Ours ($k=3$) & 32.72 & 20.00 & 20.00 & 40.00 & 40.00 & 46.67 & 26.67 & 33.33 & 20.00 & \textbf{33.33} & 46.67 & 33.33 \\
\midrule
\rowcolor{blue!8}
\multicolumn{13}{c}{\textbf{Beam depth $\gamma=2$}} \\
\midrule
Random ($k=1$) & 30.25 & 20.00 & 20.00 & 33.33 & 33.33 & 46.67 & 40.00 & 46.67 & 13.33 & 20.00 & 46.67 & 8.33 \\
Random ($k=2$) & 33.33 & \textbf{26.67} & 26.67 & 40.00 & 40.00 & 40.00 & \textbf{46.67} & 33.33 & 20.00 & 20.00 & 53.33 & 16.67 \\
Random ($k=3$) & 27.16 & \textbf{26.67} & 26.67 & 26.67 & 26.67 & 40.00 & 20.00 & 33.33 & 13.33 & 6.67 & 46.67 & 33.33 \\ \midrule
MJ ($k=1$) & 32.72 & 13.33 & 20.00 & 40.00 & 26.67 & 46.67 & 20.00 & 40.00 & 20.00 & \textbf{33.33} & 53.33 & \textbf{50.00} \\
MJ ($k=2$) & 33.95 & 6.67 & \textbf{33.33} & 26.67 & 33.33 & 40.00 & 40.00 & 53.33 & 20.00 & \textbf{33.33} & 40.00 & \textbf{50.00} \\
MJ ($k=3$) & 32.72 & 20.00 & 26.67 & 26.67 & 33.33 & 40.00 & 33.33 & \textbf{33.33} & 20.00 & 20.00 & \textbf{60.00} & 33.33 \\ \midrule
Ours ($k=1$) & 32.10 & 13.33 & 26.67 & \textbf{53.33} & 26.67 & 40.00 & 33.33 & 46.67 & 20.00 & 20.00 & 46.67 & 25.00 \\
Ours ($k=2$) & 31.48 & 6.67 & 26.67 & \textbf{53.33} & 33.33 & 33.33 & 40.00 & 60.00 & 20.00 & 20.00 & 40.00 & 8.33 \\
Ours ($k=3$) & 26.54 & 6.67 & 13.33 & 33.33 & 20.00 & 26.67 & 33.33 & 53.33 & 13.33 & 20.00 & 46.67 & 25.00 \\
\bottomrule
\end{tabular}}
\end{table}

The MMSI-Bench dataset \citep{yang2025mmsi}  covers 11 question types: Camera motion, Object motion, Positional relationship variants (Camera--Object, Object--Object, Region--Region, Camera--Camera, Camera--Region, Object--Region), Appearance attributes, Measurable attributes, and MSR (multi-step reasoning).
While the original MMSI-Bench test set contains 1,000 multiple-choice question–answer pairs based on
multiple images, we follow MindJourney \cite{Yang2025MindJourneyTS} and only use a subset of 162 questions containing at most 2 images per question. This subset has a similar size of SAT-Real \citep{ray2025sat} and includes 15 questions per category except for MSR, on which we retain all 12 questions with two images from the original test set. 

Overall, MMSI-Bench presents a substantially more challenging benchmark than SAT-Real for several key reasons. While SAT-Real features relatively straightforward binary-choice questions (e.g., "how did the camera likely rotate?" with answers "left" or "right"), MMSI-Bench employs complex four-choice questions that demand fine-grained spatial reasoning and a doubled answer space complexity. The 13 distinct question categories are further detailed when compared with  SAT-Real's 5 -- especially the six specialized positional relationship types that differentiate between  relationships, requiring models to track multiple spatial entities and their interconnections simultaneously. Moreover, MMSI-Bench questions involve more sophisticated geometric reasoning: examples include questions requiring understanding of reference frames, z-axis rotations, cardinal directions with conditional premises (e.g., "Assuming the fireplace area is on the north wall \ldots"), and multi-step spatial transformations, whereas SAT-Real questions tend toward simpler directional queries. Lastly, the dedicated MSR category  explicitly tests compositional spatial reasoning, which SAT-Real lacks.

Table \ref{tab:mmsi_reduced} shows the results of the different verifiers over MMSI-Bench for beam nodes of depth $\gamma \in \{1,2\}$. We observe that the increased difficulty of MMSI-Bench is reflected in the performance gap of the baseline InternVL3-14B which achieves only 27.16\% average accuracy (compared to 41.33\% on SAT-Real).  While we observe  improvements over the baseline InternVL3 VLM using test-time scaling, none of the verifiers show a consistent gain across increasing top-$k$ and $\gamma$ here. Namely, the random verifier with top-1 selection improves performance to 33.33\%, demonstrating that additional world model signal helps spatial reasoning on MMSI-Bench. However, increasing the beam depth to $\gamma=2$ does not consistently improve results across methods: while random top-2 with $\gamma=2$ maintains 33.33\% accuracy (matching random top-1), several MJ variants actually decrease with deeper beams (e.g., MJ (top-1) $\gamma=2$ drops to 32.72\%). Our ViSA performs competitively, with top-1 achieving 35.80\% accuracy surpassing MJ and random, though it shows more variance given that in some configurations, it underperforms the baseline InternVL3. Notably, increasing to top-3 selections  generally does not yield further improvements, suggesting that the added complexity from exploring more candidates may introduce noise or over-exploration. The positional relationship categories (especially Camera--Object and Object--Region) appear to benefit most from world model views, while MSR and certain motion categories show more limited gains.

\textbf{What do the inconsestient MMSI-Bench results imply?} We suspect that the near-random performance of all test-time verifiers on MMSI-Bench arises from a combination of information and representation bottlenecks that constrain how world-model rollouts contribute to reasoning. Unlike SAT-Real, where imagined frames enrich spatial diversity (e.g., revealing new viewpoints or motion cues), MMSI questions depend heavily on fine-grained visual or relational details—such as texture, color, or precise object alignment—that are easily degraded by the generative noise of the world model. This degradation is reflected in the lower aesthetic scores of MMSI-generated frames (4.53 overall vs.\ 5.12 on SAT; Table~\ref{tab:aesthetic_scores}), indicating reduced visual fidelity and compositional coherence. When the imagined views themselves are less photorealistic or semantically stable, verifiers like ViSA receive weaker perceptual evidence, making it difficult to distinguish genuinely helpful trajectories from noise. Moreover, since the test-time verifiers rely on high-level textual (ViSA's) or relational (MindJourney's) consistency to score imagined frames, these signals are poorly aligned with the subtle geometric or appearance shifts central to MMSI questions. As a result, even random selection performs comparably since all verifiers operate on similarly impoverished evidence. Increasing top-$k$ or beam depth amplifies this effect by adding more visually similar yet low-quality candidates, introducing redundancy rather than diversity. These findings suggest that effective test-time scaling for MMSI-like benchmarks will require higher-fidelity world models and verification objectives explicitly sensitive to fine-grained perceptual quality.

\begin{tcolorbox}[highlightstyle,
title={Our key findings}, 
label={box:bad-good-ugly}]

\vspace{1mm}

\begin{tcolorbox}[
  enhanced,
  colback=red!5,
  colframe=red!60!black,
  boxrule=0.8pt,
  arc=2mm,
  left=3mm,
  right=3mm,
  top=2mm,
  bottom=2mm,
  leftrule=3mm,
  borderline west={3mm}{0mm}{red!60!black}
]
{\large\textbf{\textcolor{red!60!black}{$\times$ The Bad:}}} 
Existing heuristic verifiers such as \textit{MindJourney} \citep{Yang2025MindJourneyTS} fail to provide meaningful guidance during test-time scaling. 
Our uncertainty-based ablations reveal that their \emph{helpfulness} scoring neither reduces uncertainty nor correlates with reasoning quality, 
and often reinforces systematic action-selection biases.
\end{tcolorbox}

\begin{tcolorbox}[
  enhanced,
  colback=green!5,
  colframe=green!60!black,
  boxrule=0.8pt,
  arc=2mm,
  left=3mm,
  right=3mm,
  top=2mm,
  bottom=2mm,
  leftrule=3mm,
  borderline west={3mm}{0mm}{green!60!black}
]
{\large\textbf{\textcolor{green!60!black}{$\checkmark$ The Good:}}} 
Our \emph{Verification through Spatial Assertions} (ViSA) introduces a principled, interpretable reward that jointly evaluates 
\emph{helpfulness} and \emph{confidence} of the verifier VLM. It improves SAT-Real accuracy by $\approx$ \textbf{10\,\%}
while yielding balanced, debiased spatial trajectories and scalable top-$k$ behavior.
\end{tcolorbox}

\begin{tcolorbox}[
  enhanced,
  colback=orange!5,
  colframe=orange!70!black,
  boxrule=0.8pt,
  arc=2mm,
  left=3mm,
  right=3mm,
  top=2mm,
  bottom=2mm,
  leftrule=3mm,
  borderline west={3mm}{0mm}{orange!70!black}
]
{\large\textbf{\textcolor{orange!70!black}{$\blacktriangle$ The Ugly:}}} 
On MMSI-Bench, none of the verifiers achieve consistent scaling. 
This exposes an \emph{information bottleneck} in current world models, whose imagined views fail to add discriminative relational evidence.
Our result highlights a fundamental limit of test-time verification when the world model itself lacks fidelity.
\end{tcolorbox}

\vspace{1mm}

\end{tcolorbox}

\subsection{Prompt template for claim generation}

\begin{prompttemplate}
\begin{lstlisting}[language=Python, basicstyle=\ttfamily\scriptsize, numbers=none]
SYSTEM PROMPT:
"You are an AI assistant that generates atomic, frame-anchored micro-claims about spatial observations when comparing two images. Your primary goal is to identify changes that help distinguish between specific answer choices for a spatial reasoning question. Generate claims that are directly relevant to the question, objectively verifiable, and useful for decision-making. Focus on binary, measurable changes rather than subjective observations. Generate EXACTLY 2-4 high-quality claims that would help a human choose between the answer options."

CONTENT STRUCTURE:
1. "After performing the action: '[action_description]'"
2. "Compare these two images and generate 2-4 specific micro-claims about what changed:"
3. "BEFORE (previous view): [input_image]"
4. "AFTER (current view): [world_model_view]"
5. "The original question is: [question]"

[If answer_choices provided:]
"Answer Choices:
  - [choice_1]
  - [choice_2]
  - [choice_n]

Focus your claims on observations that would help distinguish between these answer choices."

PRIORITIZED CHANGE TYPES:
"1. VISIBILITY CHANGES: Objects appearing/disappearing, moving in/out of frame"
"2. EDGE POSITIONING: Objects moving closer to or away from frame edges"
"3. RELATIVE POSITIONS: Clear positional shifts with reference to other objects or frame boundaries"
"4. OCCLUSION CHANGES: Objects becoming more or less hidden by other objects"

CLAIM GENERATION GUIDELINES:
"Guidelines for claim generation:"
"- Generate 2-4 claims that directly help distinguish between the answer choices"
"- If changes are minimal or unclear, generate fewer claims (quality over quantity)"
"- Each claim should be independently verifiable and distinct"
"- Prioritize claims that would help a human choose between the answer options"
"- Focus on binary, measurable changes rather than subjective observations"

HIGH-QUALITY CLAIM PATTERNS:
"Examples of HIGH-QUALITY claim patterns:"
"- [Object X] [appears/disappears] on the [left/right] side of the frame"
"- [Object X] moves [closer to/further from] the [left/right] edge"
"- [Object X] becomes [more/less] visible on the [left/right] side"

LOW-QUALITY PATTERNS TO AVOID:
"AVOID these LOW-QUALITY claim types:"
"- 'remains the same' or 'stays in position' (not useful for decision-making)"
"- Size changes ('appears larger/smaller') as these are subjective"
"- Vague movements ('shifts leftward') without clear reference points"
"- Quality assessments ('looks closer/farther') without measurable changes"
"- Multi-predicate sentences (split them into separate claims)"

FINAL REQUIREMENTS:
"Requirements:"
"- Be specific about locations and reference points"
"- Focus on binary or clear changes (visible/not visible, in frame/out of frame)"
"- Use precise directional language (left edge, right side, top corner)"
"- Make claims that would help answer the specific spatial reasoning question"


OUTPUT FORMAT:
"Output format:"
"- Output the micro-claims, one per line, starting with '- '"
"- Each claim should be a single, clear sentence"
"- Focus on the most important changes for answering the question"
\end{lstlisting}
\end{prompttemplate}

\subsection{Prompt template for claim verification}

\begin{verificationtemplate}
\begin{lstlisting}[language=Python, basicstyle=\ttfamily\scriptsize, numbers=none]
SYSTEM PROMPT:
"You are an AI assistant that verifies micro-claims against visual frames using semantic reasoning. Your task is to determine the logical relationship between a claim and the visual evidence. Focus on whether the evidence entails, contradicts, or provides insufficient information about the claim. Be precise and consider the semantic meaning of the claim in relation to what you observe. Additionally, provide a confidence score reflecting how certain you are about your judgment."

CONTENT STRUCTURE:
1. "Verify this micro-claim against the provided frames:"
2. "Claim: '[claim_text]'"
3. "Frame range: [0: input_image, 1: world_model_view]"
4. "Analyze the frames and determine the semantic relationship between the claim and evidence:"

FRAME PRESENTATION:
"For each frame in the claim's frame range:"
"Frame 1: [frame_path_1]"
"Frame 2: [frame_path_2]"


VERIFICATION INSTRUCTIONS:
"Instructions:"
"1. Examine the specific frames mentioned in the claim carefully"
"2. Determine if the visual evidence ENTAILS the claim (strongly supports it)"
"3. Check if the evidence CONTRADICTS the claim (directly opposes it)"
"4. Assess if the evidence is INSUFFICIENT (lacks information to determine support or contradiction)"
"5. Consider spatial relationships, object properties, movements, and transformations"
"6. For spatial reasoning tasks, focus on directional movements, rotations, and perspective changes"
"7. Evaluate your confidence in the judgment (0.0 = completely uncertain, 1.0 = completely certain)"
"8. Respond with the required format including verdict, confidence, and reasoning"

RESPONSE FORMAT:
"Response format:"
"VERDICT: [ENTAILED/CONTRADICTED/INSUFFICIENT]"
"CONFIDENCE: [0.0-1.0]"
"REASONING: [Clear explanation of the semantic relationship and confidence level]"

CONFIDENCE GUIDELINES (0.0-1.0 SCALE):
"- 0.95-1.0: Extremely clear, unambiguous evidence"
"- 0.85-0.94: Very clear evidence with minor uncertainties"
"- 0.70-0.84: Clear evidence with some ambiguity"
"- 0.50-0.69: Moderate evidence, noticeable uncertainty"
"- 0.30-0.49: Weak evidence, significant uncertainty"
"- 0.10-0.29: Very unclear evidence, high uncertainty"
"- 0.00-0.09: No clear evidence or contradictory signals"

IMPORTANT NOTES:
"Important: Use the full 0.0-1.0 range. Reserve 0.9+ for truly exceptional cases."

FUNCTION PARAMETERS:
- claim: dict - Contains 'text' and 'frame_range' keys
- frames: list - The input image and the world model output to analyze

OUTPUT:
Returns tuple of (system_prompt, content_list)
\end{lstlisting}
\end{verificationtemplate}

\end{document}